\documentclass[
]{ceurart}

\usepackage{listings}
\usepackage{graphicx}
\usepackage{amsmath,amssymb,amsfonts}
\usepackage{algorithmic}
\usepackage{booktabs}
\usepackage{colortbl}
\usepackage{latexsym}
\usepackage[linesnumbered,lined,boxed,commentsnumbered]{algorithm2e}

\usepackage{color}
\begin{document}

\copyrightyear{2022}
\copyrightclause{Copyright for this paper by its authors.
  Use permitted under Creative Commons License Attribution 4.0
  International (CC BY 4.0).}

\conference{NeSy 2022, 16th International Workshop on Neural-Symbolic Learning and Reasoning, Cumberland Lodge, Windsor, UK}

\title{Multi-Step Deductive Reasoning Over Natural Language: An Empirical Study on Out-of-Distribution Generalisation}


\author[1,2]{Qiming Bao}[%
email=qbao775@aucklanduni.ac.nz,
]
\cormark[1]
\address[1]{Strong AI Lab, NAOInstitute, Waipapa Taumata Rau - The University of Auckland, Auckland, New Zealand}
\address[2]{Xtracta, Auckland, New Zealand}
\address[3]{Department of Computer Science, University of Cambridge, Cambridge, The United Kingdom}

\author[1]{Alex Yuxuan Peng}[%
email=ypen260@aucklanduni.ac.nz,
]
\cormark[1]

\author[1]{Tim Hartill}[%
email=thar011@aucklanduni.ac.nz,
]

\author[1]{Neset Tan}[%
email=ntan607@aucklanduni.ac.nz,
]

\author[3]{Zhenyun Deng}[%
email=zden658@aucklanduni.ac.nz,
]

\author[1]{Michael Witbrock}[%
email=m.witbrock@auckland.ac.nz,
]
\cormark[1]

\author[1]{Jiamou Liu}[%
email=jiamou.liu@auckland.ac.nz,
]
\cormark[1]

\cortext[1]{Corresponding author.}

\begin{abstract}
Combining deep learning with symbolic logic reasoning aims to capitalize on the success of both fields and is drawing increasing attention. Inspired by DeepLogic, an end-to-end model trained to perform inference on logic programs, we introduce IMA-GloVe-GA, an iterative neural inference network for multi-step reasoning expressed in natural language. In our model, reasoning is performed using an iterative memory neural network based on RNN with a gated attention mechanism. We evaluate IMA-GloVe-GA on three datasets: PARARULES, CONCEPTRULES V1 and CONCEPTRULES V2. Experimental results show DeepLogic with gated attention can achieve higher test accuracy than DeepLogic and other RNN baseline models. Our model achieves better out-of-distribution generalisation than RoBERTa-Large when the rules have been shuffled. Furthermore, to address the issue of unbalanced distribution of reasoning depths in the current multi-step reasoning datasets, we develop PARARULE-Plus, a large dataset with more examples that require deeper reasoning steps. Experimental results show that the addition of PARARULE-Plus can increase the model's performance on examples requiring deeper reasoning depths. The source code and data are available at https://github.com/Strong-AI-Lab/Multi-Step-Deductive-Reasoning-Over-Natural-Language.
\end{abstract}


\maketitle
\vspace*{-0.8cm}
\section{Introduction}
Symbolic reasoning and deep learning remain two cornerstones in AI with profound yet divergent consequences. 
Indeed, symbolic approaches, equipped with various logic languages for knowledge representation and inference, have been the dominant paradigm in problem solving and reasoning. 
Deep learning approaches, through superior ability to capture rich semantic features from complex signals, triumph in tasks that usually require more intuitive and automatic judgements. A growing interest in AI amounts to harnessing the power from both schools, while mitigating each other's weaknesses. 
First, symbolic reasoning were suitable only when the task at hand, along with all contextual knowledge, can be encoded by rigorous and structured logic expressions, which is itself a formidable obstacle. 
Then, deep learning relies on neural networks which have not demonstrated the ability to perform iterative, multi-step reasoning, which has a gap in making them suitable tools for reasoning. Towards a trainable reasoner that is able to perform complex real-world reasoning tasks, it is important to (1) facilitate end-to-end reasoning by enabling multi-step reasoning and (2) bypass logic-based knowledge representation and make inferences directly from natural language inputs. The goal of this paper is to explore possibilities around these two objectives. We now present the research background in detail.

\begin{table}[]
\centering
\caption{\footnotesize Logic programs written in Prolog with different reasoning depths. These examples were sampled from \cite{cingillioglu2018deeplogic}.}
\label{symbolic-logic-program-prolog}
\small
\begin{tabular}{l|l|l}
\hline
1 Step            & 2 Steps           & 3 Steps           \\ \hline
$g(L,S) \ \mbox{:-} \ x(S,L).$ & $x(G,B) \ \mbox{:-} \ k(G,B).$ & $p(P,R) \ \mbox{:-} \ b(R,P).$ \\
$x(a,m).$           & $k(Z,V) \ \mbox{:-} \ g(Z,V).$ & $b(A,L) \ \mbox{:-} \ a(A,L).$ \\
$y(X) \ \mbox{:-} \ r(X).$     & $g(k,k).$           & $a(W,F) \ \mbox{:-} \ v(F,W).$ \\
$p(h).$             & $e(k,s).$           & $v(t,i).$           \\
$s(t,v).$           & $p(L,G) \ \mbox{:-} \ v(G,L).$  & $l(D) \ \mbox{:-} \ t(D).$     \\ \hline
$?$ $g(m,a).$ 1       & $?$ $x(k,k).$ 1       & $?$ $p(t,i).$ 1       \\
$?$ $g(a,m).$ 0       & $?$ $x(k,s).$ 0       & $?$ $p(i,t).$ 0       \\ \hline
\end{tabular}
\end{table}

\textbf{Logic programs}: Reasoning with logic programs is one of the key questions in AI. Here a knowledge base consists of a number of {\em rules}, i.e., (universally quantified) implications where the antecedent is a conjunction of literals and the consequent is an atom, and observed {\em facts}, i.e., ground atoms. The task would specify a {\em question} which is another ground atom and asks if the question logically follows from the knowledge base. 
Table~\ref{symbolic-logic-program-prolog} illustrates several archetypal examples of  reasoning  tasks in logic programs (as in  \cite{cingillioglu2018deeplogic}). The rules, facts, and questions are expressed in predicate logic where variables are capitalised (such as $L,S$) and constants are in small case (such as $a,m$). 
The three columns show logic programs of different reasoning {\em depths}. For example, the first column contains rule ``$g(L, S) \ \mbox{:-} \ x(S, L)$'',  fact ``$x(a,m)$'', and two questions (starting with $?$) at the bottom rows. Semantically, the rule expresses that $g(L,S)$ holds whenever $x(S,L)$ holds for any constants $L$ and $S$. From $x(a,m)$, a simple unification followed by a 1-step forward chaining inference derives $g(m,a)$, which answers the first question positively  \cite{russell2016artificial}. On the other hand,  the second column contains rules ``$x(G,B) \ \mbox{:-} \ k(G,B)$'' ``$k(Z,V) \ \mbox{:-} \ g(Z,V)$'', and fact ``$g(k,k)$''. It takes two forward chaining steps to establish  ``$x(k,k)$'' and thus has depth 2.

\smallskip

\textbf{DeepLogic}: The ability to conduct iterated inference for multi-step reasoning tasks such as the ones above is viewed as an unchallenged strength of rule-based inference algorithms. Yet recent advancements in deep learning techniques have  challenged this view. DeepLogic, introduced in \cite{cingillioglu2018deeplogic}, is an RNN-based neural network for solving reasoning tasks of logic programs. 
The model encodes logic programs at the character level and is trained on 12 different types of logic programs, without  explicitly applying any symbolic inference algorithm. 
Through a series of experiments, DeepLogic has demonstrated abilities to handle tasks that require multi-step reasoning (up to a certain small depth). 


\smallskip

\textbf{Reasoning in natural language}: The abilities demonstrated by DeepLogic has given hope for similar neural networks to perform more general reasoning tasks. In particular, neural network's key strengths involve the ability to extract rich syntactical and semantic feature from free-flowing texts, expressed in natural language. 
Indeed, PARARULES, introduced in \cite{clark2020transformers}, is a multi-step reasoning dataset expressed in natural language\footnote{{\tt https://allenai.org/data/ruletaker} \label{pararule}}. Each sample in PARARULES resembles a logic program in the style of Table~\ref{symbolic-logic-program-prolog}, except that the knowledge base (rules and facts) and questions are expressed in natural language (See Figure~\ref{pararule-example}). 
%
We thus aim to explore end-to-end neural-based multi-step reasoners over natural language using PARARULES as a testing platform, while addressing three issues: 

\smallskip

{\bf (1)} Existing models, including DeepLogic and other RNN-based baseline models, have room for improvement in terms of their reasoning abilities over natural language. The vanilla GRU/LSTM model might not handle well the multi-step reasoning tasks over logic programs from \cite{cingillioglu2018deeplogic} and natural language from Table~\ref{table1}. DeepLogic shows that with the help of GRU and dot-product attention, the model can learn to reason over logic programs from \cite{cingillioglu2018deeplogic} and natural language from Table~\ref{table1}. However, DeepLogic does not show the best performance on Table~\ref{table1}. Dynamic memory network with gated attention has shown remarkable performance on the bAbI deductive reasoning task \cite{weston2015towards}. Our first contribution is to introduce IMA-GloVe-GA, an iterative neural inference network that combined DeepLogic with gated attention, for multi-step reasoning tasks. Our model achieves the best test accuracy among the RNN-based models on the PARARULES dataset. The test accuracy of our model is on average 7.8 percentage points higher than that of DeepLogic (from Table~\ref{table1}. IMA-GloVe-GA is our model, and IMA-GloVe is from DeepLogic).

{\bf (2)} {\em Out-of-distribution (OOD) generalisation} means the test set has a distribution that is unknown or different from the distribution of the training set. In multi-step reasoning tasks, OOD generalisation means that (1) the model is able to reason for cases that have a depth greater than the depths of the instances it was trained on, and (2) the model is able to handle samples with shuffled rules from the training instances. Shuffling here means permuting the rules in the knowledge base. Being able to handle OOD is a crucial indicator of a model's reasoning capabilities. 
%
%
%
In \cite{clark2020transformers}, pretrained RoBERTa-Large \cite{liu2019roberta} 
achieves good performance on the PARARULES dataset (See Table~\ref{table1}). However, it is unclear whether the model indeed performs multi-step reasoning to the extend that it handles OOD test examples. 
Through a series of experiments, we show that RoBERTa-Large overfits and fails to generalise on examples with shuffled rules. This demonstrates that RoBERTa-Large over-utilises the ordering of rules. On the other hand, our IMA-GloVe-GA outperforms RoBERTa-Large and DeepLogic, when the models are trained on a dataset with fewer examples and unshuffled rules and are tested on a larger dataset with more relations and entities and shuffled rules (See Table~\ref{4-model-conceptrule}).

{\bf (3)} CONCEPTRULES V1 and V2 \cite{CONCEPTRULESV1,CONCEPTRULESV2} are natural-language-based multi-step reasoning datasets.  
Similar to PARARULES, the CONCEPTRULES datasets also contain samples that require deep reasoning steps (depth up to 3), and thus are suitable alternatives when evaluating models' abilities for multi-step reasoning. A common issue with all three existing datasets (PARARULES, CONCEPTRULES V1 \& V2), however, lies in their unbalanced distributions over reasoning depths. They have much fewer examples that require deep reasoning (depth $\geq$ 2) than examples that require shallow reasoning (See Table~\ref{PARARULES-AND-PARARULES-PLUS-DATA-AMOUNT}). 
To address the issue of depth imbalance, we develop a large dataset on multi-step reasoning over natural language called PARARULE-Plus that has a balanced distribution over different reasoning depths. The test accuracy on deeper depths and extra out-of-distribution examples is greatly improved when we add PARARULE-Plus in the training process (Table~\ref{pararule-different-depth-detail} and \ref{pararule-add-pararule-plus-different-depth-detail-2}). The experiment result also verifies the necessity of our dataset. 

\begin{table*}[]
\centering
\caption{\footnotesize Information about the datasets used in this paper. PARARULES has less number of examples that require deep reasoning steps. CONCEPTRULES V2 does not consider reasoning depths greater than 3. The train, dev and test set are already splitted by the author of each dataset.}
\label{PARARULES-AND-PARARULES-PLUS-DATA-AMOUNT}
\small
\resizebox{\textwidth}{!}{
\begin{tabular}{@{}llcccccc@{}}
\toprule
Dataset                      &       & Depth=0 & Depth=1 & Depth=2 & Depth=3 & Depth=4 & Depth=5 \\ \midrule
                             & Train & 290435  & 157440  & 75131   & 48010   & 9443    & 7325    \\
PARARULES                    & Dev   & 41559   & 22276   & 10833   & 6959    & 1334    & 1038    \\
                             & Test  & 83119   & 45067   & 21496   & 13741   & 2691    & 2086    \\ \midrule
                             & Train & -       & -       & 89952   & 90016   & 90010   & 90022   \\
PARARULE-Plus                & Dev   & -       & -       & 16204   & 16154   & 16150   & 16150   \\
                             & Test  & -       & -       & 2708    & 2694    & 2704    & 2692    \\ \midrule
                             & Train & 2074360 & 1310622 & 873748  & 436874  & -       & -       \\
CONCEPTRULES V2 (full)       & Dev   & 115148  & 72810   & 48540   & 24270   & -       & -       \\
                             & Test  & 115468  & 72810   & 48540   & 24270   & -       & -       \\ \midrule
                             & Train & 131646  & 74136   & 49424   & 24712   & -       & -       \\
CONCEPTRULES V2 (simplified) & Dev   & 7166    & 4116    & 2744    & 1372    & -       & -       \\
                             & Test  & 7362    & 4116    & 2744    & 1372    & -       & -       \\ \bottomrule
\end{tabular}}
\end{table*}
\section{Related Work}
Systems that integrate deep learning techniques with symbolic reasoning are called {\em neuro-symbolic systems} \cite{garcez2012neural}. Many such neural reasoning models can be viewed as logical program interpreters. Neural-symbolic machines (NSM) \cite{liang2016neural} and neural program interpreters (NPI) \cite{reed2015neural} are such examples. NSM describes a framework that consists of a seq-to-seq neural programmer, a Lisp interpreter to execute the program, and iterative maximum likelihood to train the model; NPI presents an RNN to learn and represent logic programs. NPI is designed for compositional programs, including addition, sorting, and canonicalising 3D models. Reinforcement learning has also been applied to learn Prolog-like algorithms \cite{jiang2019neural}. Distributed representations of predicates and constants for traditional symbolic reasoning engines can be learned by neural theorem provers \cite{rocktaschel2017end}. In our method, end-to-end neural networks learn representations at word level and learn to reason with natural language.

Several reasoning datasets have  been introduced for natural language-based reasoning tasks which can be used to evaluate and compare neural models' reasoning capabilities. 
Roughly speaking, the datasets can be categorised as ``shallow reasoning'' and ``deep reasoning'' tasks. The first category includes Task 15 in the bAbI dataset v1.0 \cite{weston2015towards}, conditional probes in \cite{richardson2020probing}, and ``multi-hop'' reasoning dataset HotpotQA \cite{yang2018hotpotqa}. In these datasets, the reasoning instances usually do not go beyond 2-steps. A typical example in the bAbI dataset would be ``Mouse is afraid of cats. Alice is a mouse. What is Alice afraid of? A: cats.'' and a typical example in the conditional probes of \cite{richardson2020probing} is  ``If A has visited B, then C has visited D. A visited B. Has C visited D? A: Yes.'' The main difference of HotpotQA with the other two is that the rules in HotpotQA are embedded in sentences. In HotpotQA, a sentence contains both factual information and rule information. Figure \ref{2-hop-hotpotqa} shows a 2-hop example from HotpotQA dataset.

The second category contains instances where the reasoning depth may go beyond 2. These datasets are PARARULES \cite{clark2020transformers}, CONCEPTRULES V1 \cite{CONCEPTRULESV1} and CONCEPTRULES V2 \cite{CONCEPTRULESV2}. Examples in PARARULES may require reasoning depth as deep as 5, while CONCEPTRULES V1 and V2 require depths as deep as 3. PARARULES differs from the three above in the following senses: First, solving the problems in bAbI Task 15 requires implicit rules. For example, ``Alice goes to the park. Peter goes to the restaurant. Where is Alice? A: park'' requires the rule ``A moves to B $\rightarrow$ A at B''. Contrary to the bAbI tasks, PARARULES requires reasoning with explicit rules that is more akin to logic programming. Contrary to the HotpotQA dataset, in PARARULES, the facts and rules are separate. 
%
%
One of the main issues of PARARULES, CONCEPTRULES V1 and CONCEPTRULES V2 is the unbalanced distribution over reasoning depths. The datasets have more examples of shallow reasoning (Depth$<$2) than that of deep reasoning (Depth $\geq$ 2).

Other work uses Transformer-based pretrained language models to perform tasks that require multi-step reasoning over natural language. One such work is \cite{clark2020transformers} which demonstrated that the pretrained language models (RoBERTa \cite{liu2019roberta}, and BERT \cite{devlin2018bert}) can be used for natural language-based reasoning tasks. 
However, 
it is unknown whether these pretrained language models would perform better than neural networks specifically designed for reasoning or face-checking tasks~\cite{young-etal-2022-abductionrules,tan-etal-2023-multi2claim,wang2024chatlogic,bao2022natural,doi:10.1142/S2705078521500156,tan-etal-2023-multi2claim,gendron2023large,xiao2024cora}. Attention mechanisms have been used to enable efficient inference by shortening input lengths, thereby improving text generation and classification performance~\cite{tan2023input}. Previous studies have shown that data augmentation can enhance model performance on logical reasoning tasks and knowledge-based question answering~\cite{bao2023assessing,baosymbolic,bao-etal-2024-abstract,bao2025exploring,bao2025developing,qi2023a,10.1145/3373017.3373049,Ni_Bao_Li_Qi_Denny_Warren_Witbrock_Liu_2022,young-etal-2022-abductionrules,bao2025developing}. One of our goals is to compare them against models based on iterative memory mechanisms which have been shown to perform well on multi-step reasoning tasks over logic programs. 
\begin{figure}[ht]
\small
\centering
\fbox{%
  \parbox{340pt}{%
  
(\textit{Paragraph A:}) {\color{red} \textbf{\emph{LeBron James}}} won {\color{blue} \textbf{\emph{the 2015-2016 NBA Championship}}}.

(\textit{Paragraph B:}) {\color{red} \textbf{\emph{LeBron James}}} is a basketball player for {\color{green} \emph{\textbf{Cleveland Cavaliers}}}.

(\textit{Question:}) Which team did the players who won the 2015-2016 NBA Championship play for? (\textit{Answer:}) Cleveland Cavaliers.
  }%
}
\caption{A depth-2 example from HotpotQA \cite{yang2018hotpotqa}.}	
\label{2-hop-hotpotqa}
\end{figure}

\section{Problem Definition}
We consider multi-step deductive reasoning over natural language. Each sample is a triple $(Context, Question, Answer)$ where $Context$ contains natural language implications (rules) and observations (facts) resembling a knowledge base in logic programs, $Question$ is a natural language sentence expressing an atomic fact, and $Answer\in \{\mathsf{true}, \mathsf{false}\}$ tells whether $Question$ naturally follows from $Context$.  In this regard, $(Context, Question, Answer)$ is a natural-language counterpart to a logic program. Figure~\ref{pararule-example} illustrates several examples in the PARARULES dataset \cite{clark2020transformers}. The rules are expressed in phrases such as ``If A, then B'' and ``All A are B'', and facts are represented using propositions such as ``A is happy'' and ``B is funny''. 

Multi-step reasoning requires multiple reasoning steps to answer a question. We define the reasoning {\em depths} (or {\em steps}) as the number of rules required to answer a question. For example, from ``Bob is smart.'', it takes only one rule ``All smart people are talented.'' to answer ``Bob is talented?''. This example is therefore considered as depth-1 reasoning. 
PARARULES has seven sub-datasets, each is named by the greatest depth of reasoning required to deduce which related facts support its question: depths $D = 1$, $D = 2$, $D = 3$, $D \leq 3$, $D \leq 3$  + NatLang, $D \leq 5$, $D \leq 5$ + NatLang, respectively. 
Here, NatLang means the extra out-of-distribution examples, containing about 2,000 examples. These examples include questions of different depths. They were created by paraphrasing examples using crowdsourcing. 
Crowdworkers rewrite part of the synthetic dataset using phrases such as ``often'', ``rather resembles'', and ``a bit''. For instance, sentences like ``Charlie is green, but \textit{\textbf{often}} kind'' and ``Harry \textit{\textbf{seems}} to be round'' are more natural to a human reader. 
\begin{figure}[ht]
\centering
\fbox{%
  \parbox{340pt}{%
  
(\textit{Input Facts:}) Anne is rough. Anne is blue.

(\textit{Input Rules:}) Rule 1: Cold people are rough.

Rule 2: Rough people are young.

Rule 3: If Anne is green then Anne is blue.

Rule 4: If someone is rough and nice then they are green.

Rule 5: If someone is rough and furry then they are blue.

Rule 6: All young people are cold.

Q1: Anne is cold. True/False? \textbf{[Answer: T]}

Q2: Anne is not young. True/False? \textbf{[Answer: F]}

Q3: Anne is not green. True/False? \textbf{[Answer: T]}
  }%
}
\caption{\footnotesize Examples from PARARULES \cite{clark2020transformers}. The context (facts + rules) and the question are grouped as the input, and the output is a Boolean value indicating if the question is true or false, given the context.}	
\label{pararule-example}
\end{figure}


While PARARULES separates fact and rule explicitly, CONCEPTRULES V1 and V2 \cite{CONCEPTRULESV1,CONCEPTRULESV2} put fact and rule together. 
Each of CONCEPTRULES V1 and V2 has a simplified and a full version. Both CONCEPTRULES V2 (simplified) and CONCEPTRULES V2 (full) include \textit{negation as failure (NAF)} and derivable cases. Derivable means answers can be derived from context and question. Negation as failure means if we cannot find facts or rules to derive the answer, we assume it is false. The different versions of CONCEPTULES are summarised in Table~\ref{entity-relation}. 
In CONCEPTRULES V2 (full), the rulesets are randomly shuffled and random textual noise is added. Both CONCEPTRULES V1 and V2 contain examples with reasoning depths from 0 to 3. CONCEPTRULES V1 does not label the reasoning depth for each example, while CONCEPTRULES V2 contains the labels. 
\begin{table*}[]
\centering
\caption{\footnotesize The entity types and relation types for CONCEPTRULES V1 (simplified/full), CONCEPTRULES V2 (simplified/full), PARARULES, and our PARARULE-Plus.}
\label{entity-relation}
\small
\resizebox{\textwidth}{!}{
\begin{tabular}{@{}lcccccc@{}}
\toprule
Dataset                      & \#Entity & \#Relation & Shuffled Rules & Depth Tag & Derivable & NAF \\ \midrule
CONCEPTRULES V1 (simplified) & 385      & 7          & No             & No        & Yes       & Yes \\
CONCEPTRULES V1 (full)       & 4048     & 24         & Yes            & No        & Yes       & No  \\
CONCEPTRULES V2 (simplified) & 385      & 7          & No             & Yes       & Yes       & Yes \\
CONCEPTRULES V2 (full)       & 4048     & 24         & Yes            & Yes       & Yes       & Yes \\
PARARULES                    & 19       & 4          & No             & Yes       & Yes       & Yes \\
PARARULE-Plus                & 71       & 8          & No             & Yes       & Yes       & Yes \\ \bottomrule
\end{tabular}}
\end{table*}

\section{Method}

\begin{figure}[ht]
\centering
\includegraphics[width=250pt]{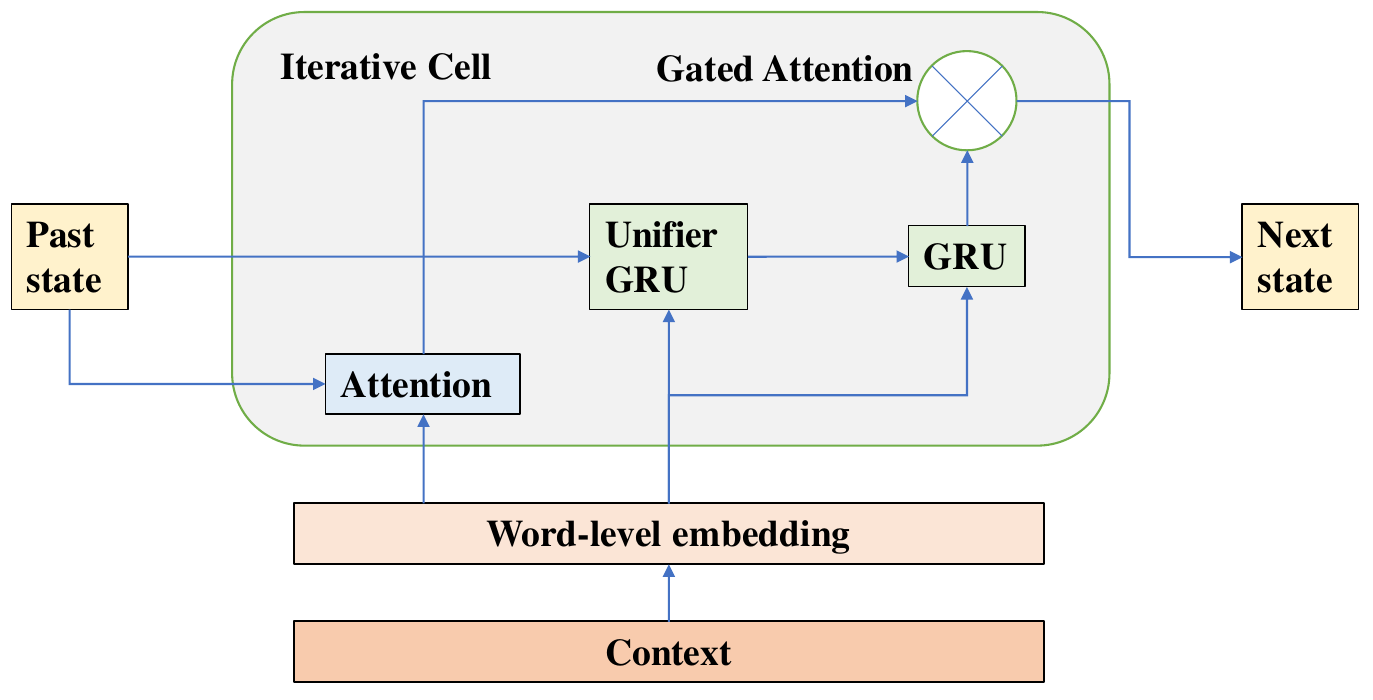}
\caption{\footnotesize The iterative neural cell with word-level embeddings as input. Questions and contexts are represented as word embeddings, and then attentions are computed to pick up related rules. Gated attention is used to compute the weighted sum of the Unifier GRU outputs. Then the weighted sum updates the state for the next iteration.}
\label{fig3}
\end{figure}

This section describes a word-level RNN-based iterative neural network. The general idea of the model is borrowed from DeepLogic. 
DeepLogic is an end-to-end iterative memory attention network trained on symbolic logic programs. For details about DeepLogic, we refer the reader to the original paper \cite{cingillioglu2018deeplogic}. The main differences in the work presented here are that we adapt the DeepLogic model to learning logic expressed in natural language, and the model operates at the word-level instead of character-level. The main architecture of the iteration framework is the same as DeepLogic (shown in Figure \ref{fig3}).


\smallskip

\noindent {\bf Word-level embedding.} The input representation layer of the network takes a sequence of words concatenated from two sentences $w_{0}^{C}, \ldots, w_{m}^{C}$ and $w_{0}^{S}, \ldots, w_{n}^{S}$ for context and question respectively. 
\begin{equation}\small h_{t}=\mathsf{G R U}\left(\mathsf{GloVe}\left[\mathbf{I}_{:: t}^{C}+\mathbf{I}_{:: t}^{S}\right], h_{t-1}\right)\label{equation1}\end{equation}
The context $\mathbf{I}^{C}$ and the question $\mathbf{I}^{S}$ at time step $t$ are embedded by $\mathsf{GloVe}[\mathbf{I}_{::t}^{C}+\mathbf{I}_{::t}^{S}]$, the GloVe \cite{pennington2014glove} word vector representation. GloVe is a set of large-scale pretrained word vectors. We use GloVe instead of the character-level embedding in DeepLogic. In DeepLogic, the logic programs are expressed by symbols using English letters and other characters. However, the logic programs in our settings are expressed in natural language. Representing the programs using word embeddings can better capture the semantic information of natural language. From the other perspective, GloVe uses ratios of co-occurrence probabilities to enlarge or narrow the relationship between words of different or similar meanings \cite{pennington2014glove}. At the same time, Word2Vec uses a local n-gram window to extract information. Furthermore, GloVe achieves better results faster than the other word-level embeddings like Word2Vec on word analogy task \cite{pennington2014glove}. The dimension of the GloVe embedding we used is $1$ $\times$ $100$. The dimension of a sentence embedding concatenated with 5 words is $5$ $\times$ $100$. The sentence embedding is processed by the gated recurrent unit (GRU) \cite{cho2014properties}. Hidden state at time $t$ is denoted as $h_t$. The context vector is $\mathbf{C} \in \mathbb{R}^{R \times L\times d}$, where $R$ is the number of rules, $L$ is the number of words in the rules and $d$ is the dimension of the embedding. 

\noindent {\bf Iteration.} Each iteration step computes the new state based on the current state and rules. In our model, gated attention is trained to decide how much of the state information will be reserved from the current state and the previous state. The process is iterated for $T$ steps ($T$ is pre-determined). The initial state is denoted as ${s}^{0} = q$, where $q$ is the question vector and $q$ $\in$ $\mathbb{R}^{d}$.  The $i$-th rule is denoted as $r_i$. $W$ and $U$ are the learnable weight matrices. $b$ is the bias vector.
\begin{equation}\small w_{i}^{t}=\left[s^{t} ; q ; r_{i} ;\left(s^{t}-r_{i}\right)^{2} ; s^{t} \odot r_{i}\right]\label{equation2}\end{equation}
\vspace*{-0.5cm}
\begin{equation}\small \alpha_{i}^{t}=\sigma(W\left(U w_{i}^{t}+b\right)+b)\label{equation3}\end{equation}

At time step $t$, we compute a feature vector $w_{i}^{t}$ using the current state $s^{t} \in \mathbb{R}^{d}$, question vector $q$ and a rule $r_i$. In \eqref{equation2}, $[;]$ is a concatenation operator. We use a feed-forward network to compute the final attention vector $\alpha_{i}^{t}$ \eqref{equation3}, where $\sigma$ is a sigmoid function. For experimental comparison, softmax is used for our IMA-GloVe-GA and another baseline model IMASM-GloVe to compute the attention vector $\alpha_{i}^{t}$, instead of the sigmoid function.
\begin{equation}\small h_{i j}^{t}=\mathsf{GRU}\left(\mathbf{C}_{i j}, h_{i(j-1)}^{t}\right)\label{equation4}\end{equation}
\begin{equation}\small s^{t+1}=\sum_{i}^{R} \alpha_{i}^{t} h_{i L}^{t}\label{equation5}\end{equation}

\noindent We use another recurrent neural network to process the context embedding $\mathbf{C}_{ij}$. The initial hidden state $h_{i0}^{t} = s^{t}$, where $s^{t}$ is the current state. For each rule, the new hidden state $h_{ij}^t$ is computed by \eqref{equation4}. In the end, the new state $s^{t + 1}$ is computed as a weighted sum of the final hidden states in \eqref{equation5}. The Unifier GRU learns the unification between pronouns (variables) and nouns (constants).
\paragraph*{\bf Gated attention.}
Dynamic memory network+ \cite{xiong2016dynamic} achieved 100\% test accuracy by using gated attention on the bAbI deductive reasoning task (Task-15). However, Task-15 is not multi-step deductive reasoning as considered in our experiment. Dynamic memory network with gated attention was shown to perform better than the one with the traditional GRU on bAbI tasks ~\cite{xiong2016dynamic}. Although gated attention can perform well on Task-15, it is worth exploring applying an iterative neural network with gated attention to multi-step deductive reasoning. This inspired us to integrate gated attention with DeepLogic which is an iterative neural inference network for multi-step reasoning. We use gated attention to replace dot-product attention, and then update the GRU utilizing the output of gated attention. Gated attention can be seen as one of variety solution from gate mechanism and attention. There are two main existing work related to gated attention. The first one is gate mechanism. 
The gate mechanism (forget gate in LSTM \cite{gers2000learning}) is designed to address the issue of forgetting in long and short term memory for recurrent neural network. The other one is gate mechanism and attention that DeepLogic shows that with the help of GRU and dot-product attention, the model's performance on the multi-step reasoning tasks can improve a lot from pure GRU without adding dot-product attention. However, in DeepLogic, the dot-product attention is added above on GRU. For each iteration, the attention is used to compute a new state based on the embeddings of the context and question with the previous time stamp state, while the gate in GRU is not specifically trained to keep the useful information with multi-step reasoning. Then we want to use the gated attention to update the gate with attention-enhanced information from context and question in order to improve multi-step reasoning. By replacing the attention vector \eqref{equation4} with gated attention $g^{t}_{i}$ \eqref{equation6}, the gated attention is used to update the internal state for GRU as in \eqref{equation7}.
\vspace*{-0.3cm}
\begin{equation}\small g_{i}^{t}=\frac{\exp \left(W\left(U w_{i}^{t}+b\right)+b\right)}{\sum_{k=1}^{R} \exp \left(W\left(U w_{k}^{t}+b\right)+b\right)}\label{equation6}\end{equation}
\vspace*{-0.4cm}
\begin{equation}\small h_{i}=g_{i}^{t} \circ \tilde{h}_{i}+\left(1-g_{i}^{t}\right) \circ h_{i-1}\label{equation7}\end{equation}
\vspace*{-1.5cm}
\section{The Datasets}\vspace*{-0.3cm}\label{section:dataset}
To investigate whether end-to-end neural networks designed for symbolic logic reasoning can be adapted to do multi-step deductive reasoning in natural language, we evaluated the models on PARARULES \cite{clark2020transformers}, CONCEPTRULES V1 \cite{CONCEPTRULESV1}, and CONCEPTRULES V2 \cite{CONCEPTRULESV2} datasets that require various depths of reasoning. The main difference between the existing datasets used to evaluate DeepLogic and the multi-step natural language reasoning datasets is that the former are based on logic programs like the examples in Table~\ref{symbolic-logic-program-prolog}, each predicate is represented by a string of symbols and characters. However, natural language is much more diverse and expressive. Both PARARULES and CONCEPTRULES are synthetically generated datasets. PARARULES also includes examples with sentences paraphrased by humans. These examples are more diverse in language and more challenging for the model. Furthermore, to address the issue of depth imbalance in the current multi-step reasoning datasets, we develop a new dataset called PARARULE-Plus, a large multi-step reasoning dataset over natural language. The dataset includes examples of four reasoning depths, from 2 to 5. There are around 100,000 samples for each depth and nearly 400,000 samples in total. The detailed information about PARARULE-Plus can be found in Table~\ref{PARARULES-AND-PARARULES-PLUS-DATA-AMOUNT} and Appendix.


\vspace*{-0.6cm}
\section{Experiments}\vspace*{-0.3cm}
We experiment with three variants of DeepLogic. The iterative memory attention (IMA) model is adopted from DeepLogic. IMASM is similar to IMA, except that IMASM uses softmax rather than sigmoid when computing attention scores. We also test three baseline models from the bAbI leaderboard: Long short-term memory (LSTM) \cite{hochreiter1997long} (The baseline method on bAbI dataset), dynamic memory networks (DMN) \cite{kumar2016ask} (100\% test accuracy on bAbI), and memory attention control networks (MAC) \cite{hudson2018compositional} (A classical method using memory network). We use GloVe \cite{pennington2014glove} as the word vector representation for the baseline methods, including IMA-GloVe, IMASM-GloVe, MAC-GloVe, DMN-GloVe, and LSTM-GloVe. Our IMA-GloVe-GA uses gated attention instead of the dot-product attention in the IMA-GloVe model.

We train the models on PARARULES, incorporating all reasoning depths (Depth=0, Depth=1, Depth= 2, Depth=3, Depth=5) and the examples paraphrased by humans (NatLang). We set the random seed to 0. We train the models using the Adam \cite{kingma2014adam} optimiser for 30 epochs. After each epoch, the data is reshuffled, and the rules in the context are reshuffled for each mini-batch. The batch size is 32. The maximum iteration depth is set to 4. The initial learning rate for Adam\footnote{Keras Optimizers, https://keras.io/api/optimizers/ \label{learning_rate}} is 1e-02. The latent dimension $d$ for GRU is set to 64. The loss function is binary cross entropy because the task is a binary classification problem. The evaluation measure is accuracy. It is computed as $\mathrm{accuracy} = n_{\mathrm{correct}} / n_{\mathrm{total}}$, where $n_{\mathrm{correct}}$ is the number of correctly classified examples and $n_{\mathrm{total}}$ is the total number of examples. We evaluate the Transformer baseline model (RoBERTa-Large) from FAIRSEQ \cite{ott2019fairseq} that was included the original paper that introduced PARARULES \cite{clark2020transformers}. We follow the official script\footnote{FAIRSEQ, https://github.com/pytorch/fairseq \label{roberta_script}} to fine-tune the model, we use the same hyperparameter in \cite{ott2019fairseq} to fine-tune RoBERTa-Large on PARARULES. We use the initial learning rate of 1e-05, and we use 16 as the batch size. We conduct all experiments using the NVIDIA 460.84 Linux Driver. The CPU version is Intel(R) Xeon(R) Gold 5218 CPU @ 2.30GHz and 16 cores. The CUDA version is 11.2 and a Quadro RTX 8000 with 48 GB GPU memory is used for our experiments.

\begin{table*}[]
\small
\centering
\caption{\footnotesize We use GloVe \cite{pennington2014glove} as the word vector representation. We use PARARULES with all depths as the training set for all models and then test them on examples with different reasoning depths (D). Comparison among our IMA-GloVe-GA, IMA-GloVe, MAC-GloVe, DMN-GloVe, IMASM-GloVe, LSTM-GloVe, and RoBERTa-Large on PARARULES test sets with different reasoning depths.}
\label{table1}
\resizebox{0.9\textwidth}{!}{
\begin{tabular}{@{}lccccccc@{}}
\toprule
Train $\downarrow$; Test $\rightarrow$   & D=1        & D=2        & D=3        & D$\leq$3 & D$\leq$3+NatLang & D$\leq$5 & D$\leq$5+NatLang \\ \midrule
IMA-GloVe     & 0.861          & 0.853          & 0.830          & 0.842              & 0.810                      & 0.792              & 0.705                      \\
MAC-GloVe     & 0.792          & 0.776          & 0.750          & 0.763              & 0.737                      & 0.701              & 0.652                      \\
DMN-GloVe     & 0.846          & 0.843          & 0.817          & 0.827              & 0.789                      & 0.779              & 0.666                      \\
IMASM-GloVe   & 0.864          & 0.855          & 0.824          & 0.838              & 0.801                      & 0.782              & 0.608                      \\
LSTM-GloVe    & 0.500          & 0.500          & 0.500          & 0.499              & 0.499                      & 0.500              & 0.500                      \\
IMA-GloVe-GA  & \textbf{0.950} & \textbf{0.943} & \textbf{0.919} & \textbf{0.927}     & \textbf{0.883}             & \textbf{0.879}     & \textbf{0.741}             \\ \midrule
RoBERTa-Large & \textbf{0.986} & \textbf{0.985} & \textbf{0.977} & \textbf{0.979}     & \textbf{0.972}             & \textbf{0.967}     & \textbf{0.949}             \\ \bottomrule
\end{tabular}}
\end{table*}

\textbf{Experimental Results}: 
Table~\ref{table1} shows the results of the RNN-based models on test sets that require different reasoning steps. Unless otherwise stated models are trained on the entire training set with all reasoning depths. The first horizontal row denotes test sets with specific reasoning depth(s), and the second horizontal row denotes the number of test samples. We find that our IMA-GloVe-GA and RoBERTa-Large achieve the 2nd and 1st best results. These two models also achieve the 2nd and 1st best results in the test set with additional human-rewritten examples (+NatLang), showing a better generalisation performance and robustness. We find that by adding gated attention the test accuracy improved over IMA-GloVe in all cases. The results support that gated attention can be more effective than dot-product attention on examples that require multi-step reasoning. IMA-GloVe and IMASM-GloVe obtained better performance than the other RNN-based baseline models. We speculate that the dot-product attention enhances the learning and representation between context and question. In addition, the vanilla LSTM fails to converge in all those cases. We get similar results as reported in DeepLogic \cite{cingillioglu2018deeplogic} that the vanilla GRU/LSTM failed on the multi-step reasoning over logic programs. A possible reason is that the vanilla LSTM is more sensitive to hyperparameter tuning. It is harder to train to converge than the other models.



The results shown in Table~\ref{table1} compare the performances of the IMA models based on the GloVe word embeddings and the pretrained RoBERTa model on the same test sets of different reasoning depths. We find that by adding gated attention the test accuracy improved over IMA-GloVe in all cases. The results support that gated attention can be more effective than dot-product attention on examples that require multi-step reasoning. We also find that the pretrained RoBERTa model achieves better results in all of the cases. A possible reason is that our model is trained from scratch without any large-scale pretraining. However, our model can perform better than the other baselines without any pretraining.

\begin{table*}[]
\centering
\caption{\footnotesize IMA-GloVe, IMA-GloVe-GA, and RoBERTa-Large trained on CONCEPTRULES V1 (simplified / full) and tested on different test sets. Rules in CONCEPTRULES V1 Simplified are not shuffled, while CONCEPTRULES V1 full contains randomly shuffled rules. CONCEPTRULES V1 full has larger number of relations and entities than CONCEPTRULES V1 simplified.}
\label{4-model-conceptrule}
\small
\begin{tabular}{lccc}
\hline
Model                               & Train set    & \begin{tabular}[c]{@{}c@{}}Test accuracy\\ (Simplified Test set)\end{tabular} & \begin{tabular}[c]{@{}c@{}}Test accuracy\\ (Full Test set)\end{tabular} \\ \hline
IMA-GloVe          & Simplified & 0.994                                                                         & 0.729                                                                                                                                           \\
                                    & Full       & 0.844                                                                         & \textbf{0.997}                                                                                                                         \\ \hline
IMA-GloVe-GA         & Simplified & \textbf{0.998}                                                                & \textbf{0.747}                                                                                                                                   \\
                                    & Full       & 0.851                                                                & \textbf{0.999}                                                                                                                        \\ \hline
RoBERTa-Large      & Simplified & \textbf{0.997}                                                                         & 0.503                                                                                                        \\
                                    & Full       & \textbf{0.927}                                                                         & 0.995                                                                                                                   \\ \hline
\end{tabular}
\end{table*}
Table~\ref{4-model-conceptrule} shows the results on CONCEPTRULES V1 (simplified) and CONCEPTRULES V1 (full). The second column is the training set CONCEPTRULES V1 (simplified or full). The last two columns show test accuracy on CONCEPTRULES V1 (simplified or full) test set. For example, in the first row, we train IMA-GloVe on the simplified version of the training sets, and then test on different test sets. We find that IMA-GloVe-GA (with gated attention) performs better than IMA-GloVe in all cases. In contrast, the pretrained RoBERTa model only achieves high performance in the simplified test set when it is trained on the simplified training set. It indicates that the pretrained RoBERTa overfits CONCEPTRULES V1 (simplified) since the rules are not shuffled. Most likely it learns spurious relations in the training data and fails to generalise on test examples with shuffled rules. 

\begin{table*}[]
\centering
\caption{\footnotesize IMA-GloVe, IMA-GloVe-GA, and RoBERTa-Large trained on CONCEPTRULES V2 (full) and tested on test sets that require different depths of reasoning.}
\label{conceptrulev2-full-detail}
\small
\begin{tabular}{llcccccc}
\hline
Model              & Test set & \begin{tabular}[c]{@{}c@{}}Mod1\\ Depth=1\end{tabular} & \begin{tabular}[c]{@{}c@{}}Mod2\\ Depth=2\end{tabular} & \begin{tabular}[c]{@{}c@{}}Mod3\\ Depth=3\end{tabular} & \begin{tabular}[c]{@{}c@{}}Mod01\\ Depth$\leq$1\end{tabular} & \begin{tabular}[c]{@{}c@{}}Mod012\\ Depth$\leq$2\end{tabular} & \begin{tabular}[c]{@{}c@{}}Mod0123\\ Depth$\leq$3\end{tabular} \\ \hline
                   & Depth=1  & \textbf{0.999}                                         & \textbf{0.998}                                                  & \textbf{0.990}                                                  & \textbf{0.997}                                                     & \textbf{0.998}                                                      & \textbf{0.997}                                                       \\
IMA-GloVe         & Depth=2  & \textbf{0.998}                                                  & \textbf{0.999}                                         & \textbf{0.988}                                                  & \textbf{0.995}                                                              & \textbf{0.998}                                                      & \textbf{0.997}                                                       \\
                   & Depth=3  & \textbf{0.997}                                                  & 0.998                                                  & 0.981                                         & \textbf{0.991}                                                              & 0.996                                                               & \textbf{0.997}                                                       \\ \hline
                   & Depth=1  & 0.993                                                  & 0.996                                                  & 0.987                                                  & 0.987                                                              & 0.991                                                               & \textbf{0.997}                                                       \\
IMA-GloVe-GA     & Depth=2  & 0.993                                                  & \textbf{0.999}                                         & 0.974                                                  & 0.986                                                     & 0.991                                                               & 0.995                                                                \\
                   & Depth=3  & 0.988                                                  & \textbf{1}                                             & \textbf{0.994}                                         & 0.989                                                              & \textbf{0.997}                                                      & 0.994                                                                \\ \hline
                   & Depth=1  & 0.998                                                  & 0.975                                                  & 0.831                                                  & 0.995                                                              & 0.975                                                               & 0.971                                                                \\
RoBERTa-Large      & Depth=2  & 0.997                                                  & 0.972                                                  & 0.885                                                  & 0.993                                                              & 0.972                                                               & 0.965                                                                \\
                   & Depth=3  & 0.987                                                  & 0.951                                                  & 0.984                                                  & 0.988                                                              & 0.951                                                               & 0.936                                                                \\ \hline
\end{tabular}
\end{table*}
Table~\ref{conceptrulev2-full-detail} shows results on CONCEPTRULES V2. We select the four models used in Table~\ref{4-model-conceptrule}, and respectively select data of different depths in CONCEPTRULES V2 (full) as the training data. For example, Mod3 represents a model trained on examples with depth of 3. Mod0123 represents a model trained on examples with depth of 3 and less. Each row of the table represents a test set of a specific reasoning depth. We find that the IMA-based model achieves better results in almost all cases. The pretrained RoBERTa model does not outperform the IMA-based models in all cases. We find that when the model is only trained on examples with reasoning step of 1, the test accuracy dropped as we increase the reasoning steps in the test set. However, the drop in performance is not as great as we expected for the IMA-based models. This shows that the IMA-based models generalise better on out-of-distribution test examples. A possible reason why models achieve higher test accuracies on CONCEPTRULES V2 (full) is that CONCEPTRULES V2 has much more training data compared to PARARULES, from Table~\ref{PARARULES-AND-PARARULES-PLUS-DATA-AMOUNT}.


\begin{table*}[]
\centering
\caption{\footnotesize RoBERTa-Large trained on PARARULES with different reasoning depths and tested on test sets that require different depths of reasoning. A bold number indicates the highest accuracy in a test set.}
\label{pararule-different-depth-detail}
\small
\begin{tabular}{@{}lccccc@{}}
\toprule
Model         & Test set                   & \begin{tabular}[c]{@{}c@{}}Mod012\\ (Depth$\leq$2)\end{tabular} & \begin{tabular}[c]{@{}c@{}}Mod0123\\ (Depth$\leq$3)\end{tabular} & \begin{tabular}[c]{@{}c@{}}Mod0123Nat\\ (Depth$\leq$3+NatLang)\end{tabular} & \begin{tabular}[c]{@{}c@{}}Mod012345\\ (Depth$\leq$5)\end{tabular} \\ \midrule
              & Depth=0                    & \textbf{0.971}                                                                & 0.946                                                                 & 0.968                                                                                                                                             & 0.953                                                                   \\
              & Depth=1                    & \textbf{0.943}                                                                & 0.907                                                                 & 0.933                                                                                                                                             & 0.909                                                                   \\
              & Depth=2                    & \textbf{0.933}                                                                & 0.902                                                                 & 0.932                                                                                                                                             & 0.902                                                                   \\
RoBERTa-Large & Depth=3                    & 0.562                                                                & 0.902                                                                 & \textbf{0.926}                                                                                                                                             & 0.907                                                                   \\
              & Depth=4                    & 0.481                                                                & 0.863                                                                 & \textbf{0.904}                                                                                                                                              & 0.888                                                                   \\
              & Depth=5                    & 0.452                                                                & 0.856                                                                 & 0.916                                                                                                                                             & \textbf{0.933}                                                                   \\
              & NatLang & 0.573                                                                   & 0.579                                                                    & \textbf{0.962}                                                                                                                                                & 0.594                                                                   \\ \bottomrule
\end{tabular}
\end{table*}
Table~\ref{pararule-different-depth-detail} shows the results of fine-tuning RoBERTa-Large on datasets with different reasoning depths and testing on test sets of various reasoning depths. Depth$\leq$2 means the model is trained on the dataset with depths no larger than 2. Depth$\leq$3+NatLang represents the model that is trained on the dataset with depths no larger than 3 and extra examples paraphrase by humans. The models trained on the datasets with shallow depths have lower test accuracies on test sets that require deeper reasoning depths. We find that adding training examples that require deeper reasoning steps for Mod0123 improves the results on the test sets with deeper reasoning steps (e.g. Mod0123Nat vs. Mod0123, and Mod012345 vs. Mod0123). Additionally, we find that adding the examples paraphrased by humans improves the performance on test examples with deep reasoning steps and human-paraphrased test examples.

\begin{table*}[]
\centering
\caption{\footnotesize RoBERTa-Large is fine-tuned on examples with different depths from PARARULES and also the entire PARARULE-Plus(PPT), and then is evaluated on test sets that require different depths of reasoning. The yellow background indicates improvement on accuracy after adding our PARARULE-Plus in the training process.}
\label{pararule-add-pararule-plus-different-depth-detail-2}
\small
\resizebox{\textwidth}{!}{
\begin{tabular}{@{}lccccc@{}}
\toprule
Model         & Test set                   & \begin{tabular}[c]{@{}c@{}}Mod012\\ (Depth$\leq$2+PPT)\end{tabular} & \begin{tabular}[c]{@{}c@{}}Mod0123\\ (Depth$\leq$3+PPT)\end{tabular} & \begin{tabular}[c]{@{}c@{}}Mod0123Nat\\ (Depth$\leq$3+NatLang+PPT)\end{tabular} & \begin{tabular}[c]{@{}c@{}}Mod012345\\ (Depth$\leq$5+PPT)\end{tabular} \\ \midrule
              & Depth=0                    & 0.946                                                                                  & 0.901                                                                                   & 0.965                                                                                                                                                                                    & \cellcolor{yellow}\textbf{0.963 (+0.010)}                                                                                        \\
              & Depth=1                    & 0.877                                                                                  & 0.847                                                                                   & \cellcolor{yellow}\textbf{0.937 (+0.004)}                                                                                                                                                                                    & 0.881                                                                                        \\
              & Depth=2                    & 0.868                                                                                  & 0.873                                                                                   & \textbf{0.927}                                                                                                                                                                                     & 0.839                                                                                        \\
RoBERTa-Large & Depth=3                    & \cellcolor{yellow}0.771 (+0.209)                                                                                  & 0.862                                                                                   & \textbf{0.904}                                                                                                                                                                                     & 0.826                                                                                        \\
              & Depth=4                    & \cellcolor{yellow}0.675 (+0.194)                                                                                  & 0.852                                                                                   & \textbf{0.897}                                                                                                                                                                                      & 0.832                                                                                        \\
              & Depth=5                    & \cellcolor{yellow}0.661 (+0.209)                                                                                & \cellcolor{yellow}0.888 (+0.032)                                                                                   & \cellcolor{yellow}0.923 (+0.007)                                                                                                                                                                                   & \cellcolor{yellow}\textbf{0.934 (+0.001)}                                                                                        \\
              & NatLang & 0.557                                                                                  & \cellcolor{yellow}0.593 (+0.014)                                                                                  & \cellcolor{yellow}\textbf{0.970 (+0.008)}                                                                                                                                                                                     & \cellcolor{yellow}0.649 (+0.055)                                                                                        \\ \bottomrule
\end{tabular}}
\end{table*}

Table~\ref{pararule-add-pararule-plus-different-depth-detail-2} shows the results of fine-tuning RoBERTa-Large on the datasets found in Table~\ref{pararule-different-depth-detail} and our PARARULE-Plus. The addition of PARARULE-Plus during the fine-tuning improves the performance on the examples that require more reasoning steps. The yellow background shows the improvement over the test accuracy reported in Table~\ref{pararule-different-depth-detail}, and the bracket contains the magnitude of the improvement. The bold numbers indicate the highest test accuracy on corresponding test sets. The results support that our PARARULE-Plus addresses the depth imbalance issue of the current datasets and the addition of it during training improves the model's generalisation on the examples that require deeper reasoning steps.

\vspace*{-0.6cm}
\section{Conclusion}\vspace*{-0.3cm}
We provide insights into an RNN-based iterative memory model that incorporates gated attention on multi-step reasoning over natural language. Instead of  using the original GRU and dot-product attention, we integrate gated attention to update hidden states. The experiment results show the model with gated attention achieves generally better performance than the original RNN-based iterative-memory model with dot-product attention and other RNN-based models. The performance of our model is comparable or better than the much larger and pretrained RoBERTa-Large in some scenarios. Furthermore, our model shows better out-of-distribution generalisation performance than the pretained RoBERTa. 
To address the issue of depth-imbalance in the existing datasets on multi-step reasoning over natural language, we develop a large-scale multi-step reasoning dataset called PARARULE-Plus, with more examples of deep reasoning depths than previous datasets. We find that the performance of the models in our experiments improves when we add PARARULE-Plus in the training, especially on examples that require deeper reasoning depths and extra out-of-distribution examples. 


\bibliography{custom}

\begin{thebibliography}{40}
\expandafter\ifx\csname natexlab\endcsname\relax\def\natexlab#1{#1}\fi
\providecommand{\url}[1]{\texttt{#1}}
\providecommand{\href}[2]{#2}
\providecommand{\path}[1]{#1}
\providecommand{\DOIprefix}{doi:}
\providecommand{\ArXivprefix}{arXiv:}
\providecommand{\URLprefix}{URL: }
\providecommand{\Pubmedprefix}{pmid:}
\providecommand{\doi}[1]{\href{http://dx.doi.org/#1}{\path{#1}}}
\providecommand{\Pubmed}[1]{\href{pmid:#1}{\path{#1}}}
\providecommand{\bibinfo}[2]{#2}
\ifx\xfnm\relax \def\xfnm[#1]{\unskip,\space#1}\fi
\bibitem[{Cingillioglu and Russo(2019)}]{cingillioglu2018deeplogic}
\bibinfo{author}{N.~Cingillioglu}, \bibinfo{author}{A.~Russo},
\newblock \bibinfo{title}{Deeplogic: Towards end-to-end differentiable logical
  reasoning},
\newblock in: \bibinfo{booktitle}{AAAI-MAKE}, \bibinfo{year}{2019}.
\bibitem[{Russell et~al.(2016)Russell, Norvig, and
  Davis}]{russell2016artificial}
\bibinfo{author}{S.~Russell}, \bibinfo{author}{P.~Norvig},
  \bibinfo{author}{E.~Davis}, \bibinfo{title}{Artificial intelligence: A modern
  approach (3rd (global) ed.)}, \bibinfo{year}{2016}.
\bibitem[{Clark et~al.(2020)Clark, Tafjord, and
  Richardson}]{clark2020transformers}
\bibinfo{author}{P.~Clark}, \bibinfo{author}{O.~Tafjord},
  \bibinfo{author}{K.~Richardson},
\newblock \bibinfo{title}{Transformers as soft reasoners over language},
\newblock in: \bibinfo{booktitle}{IJCAI}, \bibinfo{year}{2020}, pp.
  \bibinfo{pages}{3882--3890}.
\bibitem[{Weston et~al.(2016)Weston, Bordes, Chopra, and
  Mikolov}]{weston2015towards}
\bibinfo{author}{J.~Weston}, \bibinfo{author}{A.~Bordes},
  \bibinfo{author}{S.~Chopra}, \bibinfo{author}{T.~Mikolov},
\newblock \bibinfo{title}{Towards ai-complete question answering: {A} set of
  prerequisite toy tasks},
\newblock in: \bibinfo{booktitle}{ICLR}, \bibinfo{year}{2016}.
\bibitem[{Liu et~al.(2019)Liu, Ott, Goyal, Du, Joshi, Chen, Levy, Lewis,
  Zettlemoyer, and Stoyanov}]{liu2019roberta}
\bibinfo{author}{Y.~Liu}, \bibinfo{author}{M.~Ott}, \bibinfo{author}{N.~Goyal},
  \bibinfo{author}{J.~Du}, \bibinfo{author}{M.~Joshi},
  \bibinfo{author}{D.~Chen}, \bibinfo{author}{O.~Levy},
  \bibinfo{author}{M.~Lewis}, \bibinfo{author}{L.~Zettlemoyer},
  \bibinfo{author}{V.~Stoyanov},
\newblock \bibinfo{title}{Roberta: A robustly optimized bert pretraining
  approach},
\newblock \bibinfo{journal}{arXiv preprint arXiv:1907.11692}
  (\bibinfo{year}{2019}).
\bibitem[{Hartill(2020{\natexlab{a}})}]{CONCEPTRULESV1}
\bibinfo{author}{T.~Hartill}, \bibinfo{title}{Conceptrules {V}1 dataset},
  \bibinfo{howpublished}{https://bit.ly/3uVemXG},
  \bibinfo{year}{2020}{\natexlab{a}}.
\bibitem[{Hartill(2020{\natexlab{b}})}]{CONCEPTRULESV2}
\bibinfo{author}{T.~Hartill}, \bibinfo{title}{Conceptrules {V}2 dataset},
  \bibinfo{howpublished}{https://bit.ly/3PApIIB},
  \bibinfo{year}{2020}{\natexlab{b}}.
\bibitem[{Garcez et~al.(2012)Garcez, Broda, and Gabbay}]{garcez2012neural}
\bibinfo{author}{A.~S.~d. Garcez}, \bibinfo{author}{K.~B. Broda},
  \bibinfo{author}{D.~M. Gabbay}, \bibinfo{title}{Neural-symbolic learning
  systems: foundations and applications}, \bibinfo{year}{2012}.
\bibitem[{Liang et~al.(2017)Liang, Berant, Le, Forbus, and
  Lao}]{liang2016neural}
\bibinfo{author}{C.~Liang}, \bibinfo{author}{J.~Berant},
  \bibinfo{author}{Q.~Le}, \bibinfo{author}{K.~D. Forbus},
  \bibinfo{author}{N.~Lao},
\newblock \bibinfo{title}{Neural symbolic machines: Learning semantic parsers
  on {F}reebase with weak supervision},
\newblock in: \bibinfo{booktitle}{ACL}, \bibinfo{year}{2017}, pp.
  \bibinfo{pages}{23--33}.
\bibitem[{Reed and De~Freitas(2015)}]{reed2015neural}
\bibinfo{author}{S.~Reed}, \bibinfo{author}{N.~De~Freitas},
\newblock \bibinfo{title}{Neural programmer-interpreters},
\newblock \bibinfo{journal}{arXiv preprint arXiv:1511.06279}
  (\bibinfo{year}{2015}).
\bibitem[{Jiang and Luo(2019)}]{jiang2019neural}
\bibinfo{author}{Z.~Jiang}, \bibinfo{author}{S.~Luo},
\newblock \bibinfo{title}{Neural logic reinforcement learning},
\newblock in: \bibinfo{booktitle}{ICML}, \bibinfo{organization}{PMLR},
  \bibinfo{year}{2019}, pp. \bibinfo{pages}{3110--3119}.
\bibitem[{Rockt{\"{a}}schel and Riedel(2017)}]{rocktaschel2017end}
\bibinfo{author}{T.~Rockt{\"{a}}schel}, \bibinfo{author}{S.~Riedel},
\newblock \bibinfo{title}{End-to-end differentiable proving},
\newblock in: \bibinfo{booktitle}{NIPS}, \bibinfo{year}{2017}, pp.
  \bibinfo{pages}{3788--3800}.
\bibitem[{Richardson et~al.(2020)Richardson, Hu, Moss, and
  Sabharwal}]{richardson2020probing}
\bibinfo{author}{K.~Richardson}, \bibinfo{author}{H.~Hu},
  \bibinfo{author}{L.~S. Moss}, \bibinfo{author}{A.~Sabharwal},
\newblock \bibinfo{title}{Probing natural language inference models through
  semantic fragments.},
\newblock in: \bibinfo{booktitle}{AAAI}, \bibinfo{year}{2020}, pp.
  \bibinfo{pages}{8713--8721}.
\bibitem[{Yang et~al.(2018)Yang, Qi, Zhang, Bengio, Cohen, Salakhutdinov, and
  Manning}]{yang2018hotpotqa}
\bibinfo{author}{Z.~Yang}, \bibinfo{author}{P.~Qi}, \bibinfo{author}{S.~Zhang},
  \bibinfo{author}{Y.~Bengio}, \bibinfo{author}{W.~W. Cohen},
  \bibinfo{author}{R.~Salakhutdinov}, \bibinfo{author}{C.~D. Manning},
\newblock \bibinfo{title}{Hotpotqa: A dataset for diverse, explainable
  multi-hop question answering},
\newblock \bibinfo{journal}{arXiv preprint arXiv:1809.09600}
  (\bibinfo{year}{2018}).
\bibitem[{Devlin et~al.(2019)Devlin, Chang, Lee, and
  Toutanova}]{devlin2018bert}
\bibinfo{author}{J.~Devlin}, \bibinfo{author}{M.-W. Chang},
  \bibinfo{author}{K.~Lee}, \bibinfo{author}{K.~Toutanova},
\newblock \bibinfo{title}{{BERT}: Pre-training of deep bidirectional
  transformers for language understanding},
\newblock in: \bibinfo{booktitle}{NAACL-HLT}, \bibinfo{year}{2019}.
\bibitem[{Young et~al.(2022)Young, Bao, Bensemann, and
  Witbrock}]{young-etal-2022-abductionrules}
\bibinfo{author}{N.~Young}, \bibinfo{author}{Q.~Bao},
  \bibinfo{author}{J.~Bensemann}, \bibinfo{author}{M.~Witbrock},
\newblock \bibinfo{title}{{A}bduction{R}ules: Training transformers to explain
  unexpected inputs},
\newblock in: \bibinfo{editor}{S.~Muresan}, \bibinfo{editor}{P.~Nakov},
  \bibinfo{editor}{A.~Villavicencio} (Eds.), \bibinfo{booktitle}{Findings of
  the Association for Computational Linguistics: ACL 2022},
  \bibinfo{publisher}{Association for Computational Linguistics},
  \bibinfo{address}{Dublin, Ireland}, \bibinfo{year}{2022}, pp.
  \bibinfo{pages}{218--227}. \URLprefix
  \url{https://aclanthology.org/2022.findings-acl.19/}.
  \DOIprefix\doi{10.18653/v1/2022.findings-acl.19}.
\bibitem[{Tan et~al.(2023)Tan, Nguyen, Bensemann, Peng, Bao, Chen, Gahegan, and
  Witbrock}]{tan-etal-2023-multi2claim}
\bibinfo{author}{N.~Tan}, \bibinfo{author}{T.~Nguyen},
  \bibinfo{author}{J.~Bensemann}, \bibinfo{author}{A.~Peng},
  \bibinfo{author}{Q.~Bao}, \bibinfo{author}{Y.~Chen},
  \bibinfo{author}{M.~Gahegan}, \bibinfo{author}{M.~Witbrock},
\newblock \bibinfo{title}{{M}ulti2{C}laim: Generating scientific claims from
  multi-choice questions for scientific fact-checking},
\newblock in: \bibinfo{editor}{A.~Vlachos}, \bibinfo{editor}{I.~Augenstein}
  (Eds.), \bibinfo{booktitle}{Proceedings of the 17th Conference of the
  European Chapter of the Association for Computational Linguistics},
  \bibinfo{publisher}{Association for Computational Linguistics},
  \bibinfo{address}{Dubrovnik, Croatia}, \bibinfo{year}{2023}, pp.
  \bibinfo{pages}{2652--2664}. \URLprefix
  \url{https://aclanthology.org/2023.eacl-main.194/}.
  \DOIprefix\doi{10.18653/v1/2023.eacl-main.194}.
\bibitem[{Wang et~al.(2024)Wang, Liu, Bao, Rong, and Zhang}]{wang2024chatlogic}
\bibinfo{author}{Z.~Wang}, \bibinfo{author}{J.~Liu}, \bibinfo{author}{Q.~Bao},
  \bibinfo{author}{H.~Rong}, \bibinfo{author}{J.~Zhang},
\newblock \bibinfo{title}{Chatlogic: Integrating logic programming with large
  language models for multi-step reasoning},
\newblock in: \bibinfo{booktitle}{2024 International Joint Conference on Neural
  Networks (IJCNN)}, \bibinfo{organization}{IEEE}, \bibinfo{year}{2024}, pp.
  \bibinfo{pages}{1--8}.
\bibitem[{Bao et~al.(2022)Bao, Witbrock, and Liu}]{bao2022natural}
\bibinfo{author}{Q.~Bao}, \bibinfo{author}{M.~Witbrock},
  \bibinfo{author}{J.~Liu},
\newblock \bibinfo{title}{Natural language processing and reasoning},
\newblock \bibinfo{journal}{IEEE Vehicular Technology Society Invited Talk}
  (\bibinfo{year}{2022}).
\bibitem[{Bensemann et~al.(2022)Bensemann, Bao, Gendron, Hartill, and
  Witbrock}]{doi:10.1142/S2705078521500156}
\bibinfo{author}{J.~Bensemann}, \bibinfo{author}{Q.~Bao},
  \bibinfo{author}{G.~Gendron}, \bibinfo{author}{T.~Hartill},
  \bibinfo{author}{M.~Witbrock},
\newblock \bibinfo{title}{Relating blindsight and ai: A review},
\newblock \bibinfo{journal}{Journal of Artificial Intelligence and
  Consciousness} \bibinfo{volume}{09} (\bibinfo{year}{2022})
  \bibinfo{pages}{111--125}. \URLprefix
  \url{https://doi.org/10.1142/S2705078521500156}.
  \DOIprefix\doi{10.1142/S2705078521500156}.
  \href{http://arxiv.org/abs/https://doi.org/10.1142/S2705078521500156}{{\tt
  arXiv:https://doi.org/10.1142/S2705078521500156}}.
\bibitem[{Gendron et~al.(2023)Gendron, Bao, Witbrock, and
  Dobbie}]{gendron2023large}
\bibinfo{author}{G.~Gendron}, \bibinfo{author}{Q.~Bao},
  \bibinfo{author}{M.~Witbrock}, \bibinfo{author}{G.~Dobbie},
\newblock \bibinfo{title}{Large language models are not strong abstract
  reasoners},
\newblock \bibinfo{journal}{arXiv preprint arXiv:2305.19555}
  (\bibinfo{year}{2023}).
\bibitem[{Xiao et~al.(2024)Xiao, Shen, Bao, Rong, Liu, Wang, and
  Liu}]{xiao2024cora}
\bibinfo{author}{X.~Xiao}, \bibinfo{author}{S.~Shen}, \bibinfo{author}{Q.~Bao},
  \bibinfo{author}{H.~Rong}, \bibinfo{author}{K.~Liu},
  \bibinfo{author}{Z.~Wang}, \bibinfo{author}{J.~Liu},
\newblock \bibinfo{title}{Cora: Optimizing low-rank adaptation with common
  subspace of large language models},
\newblock \bibinfo{journal}{arXiv preprint arXiv:2409.02119}
  (\bibinfo{year}{2024}).
\bibitem[{Tan et~al.(2023)Tan, Peng, Bensemann, Bao, Hartill, Gahegan, and
  Witbrock}]{tan2023input}
\bibinfo{author}{N.~{\"O}. Tan}, \bibinfo{author}{A.~Y. Peng},
  \bibinfo{author}{J.~Bensemann}, \bibinfo{author}{Q.~Bao},
  \bibinfo{author}{T.~Hartill}, \bibinfo{author}{M.~Gahegan},
  \bibinfo{author}{M.~Witbrock},
\newblock \bibinfo{title}{Input-length-shortening and text generation via
  attention values},
\newblock \bibinfo{journal}{arXiv preprint arXiv:2303.07585}
  (\bibinfo{year}{2023}).
\bibitem[{Bao et~al.(2023)Bao, Gendron, Peng, Zhong, Tan, Chen, Witbrock, and
  Liu}]{bao2023assessing}
\bibinfo{author}{Q.~Bao}, \bibinfo{author}{G.~Gendron}, \bibinfo{author}{A.~Y.
  Peng}, \bibinfo{author}{W.~Zhong}, \bibinfo{author}{N.~Tan},
  \bibinfo{author}{Y.~Chen}, \bibinfo{author}{M.~Witbrock},
  \bibinfo{author}{J.~Liu},
\newblock \bibinfo{title}{Assessing and enhancing the robustness of large
  language models with task structure variations for logical reasoning},
\newblock \bibinfo{journal}{arXiv preprint arXiv:2310.09430}
  (\bibinfo{year}{2023}).
\bibitem[{Bao et~al.(2021)Bao, Witbrock, and Liu}]{baosymbolic}
\bibinfo{author}{Q.~Bao}, \bibinfo{author}{M.~Witbrock},
  \bibinfo{author}{J.~Liu},
\newblock \bibinfo{title}{From symbolic logic reasoning to soft reason-ing: A
  neural-symbolic paradigm},
\newblock \bibinfo{journal}{The 1st New Zealand Workshop on Artificial
  Intelligence Research (NZAIR)}  (\bibinfo{year}{2021}).
\bibitem[{Bao et~al.(2024)Bao, Peng, Deng, Zhong, Gendron, Pistotti, Tan,
  Young, Chen, Zhu, Denny, Witbrock, and Liu}]{bao-etal-2024-abstract}
\bibinfo{author}{Q.~Bao}, \bibinfo{author}{A.~Peng}, \bibinfo{author}{Z.~Deng},
  \bibinfo{author}{W.~Zhong}, \bibinfo{author}{G.~Gendron},
  \bibinfo{author}{T.~Pistotti}, \bibinfo{author}{N.~Tan},
  \bibinfo{author}{N.~Young}, \bibinfo{author}{Y.~Chen},
  \bibinfo{author}{Y.~Zhu}, \bibinfo{author}{P.~Denny},
  \bibinfo{author}{M.~Witbrock}, \bibinfo{author}{J.~Liu},
\newblock \bibinfo{title}{{A}bstract {M}eaning {R}epresentation-based
  logic-driven data augmentation for logical reasoning},
\newblock in: \bibinfo{editor}{L.-W. Ku}, \bibinfo{editor}{A.~Martins},
  \bibinfo{editor}{V.~Srikumar} (Eds.), \bibinfo{booktitle}{Findings of the
  Association for Computational Linguistics: ACL 2024},
  \bibinfo{publisher}{Association for Computational Linguistics},
  \bibinfo{address}{Bangkok, Thailand}, \bibinfo{year}{2024}, pp.
  \bibinfo{pages}{5914--5934}. \URLprefix
  \url{https://aclanthology.org/2024.findings-acl.353/}.
  \DOIprefix\doi{10.18653/v1/2024.findings-acl.353}.
\bibitem[{Bao et~al.(2025)Bao, Leinonen, Peng, Zhong, Gendron, Pistotti, Huang,
  Denny, Witbrock, and Liu}]{bao2025exploring}
\bibinfo{author}{Q.~Bao}, \bibinfo{author}{J.~Leinonen}, \bibinfo{author}{A.~Y.
  Peng}, \bibinfo{author}{W.~Zhong}, \bibinfo{author}{G.~Gendron},
  \bibinfo{author}{T.~Pistotti}, \bibinfo{author}{A.~Huang},
  \bibinfo{author}{P.~Denny}, \bibinfo{author}{M.~Witbrock},
  \bibinfo{author}{J.~Liu},
\newblock \bibinfo{title}{Exploring iterative enhancement for improving
  learnersourced multiple-choice question explanations with large language
  models},
\newblock in: \bibinfo{booktitle}{Proceedings of the AAAI Conference on
  Artificial Intelligence}, volume~\bibinfo{volume}{39}, \bibinfo{year}{2025},
  pp. \bibinfo{pages}{28955--28963}.
\bibitem[{Bao(2025)}]{bao2025developing}
\bibinfo{author}{Q.~Bao}, \bibinfo{title}{Developing And Assessing Language
  Models For Logical Reasoning Over Natural Language}, Ph.D. thesis, University
  of Auckland, \bibinfo{year}{2025}.
\bibitem[{Qi et~al.(2023)Qi, Bao, Peng, Liu, and Witbrock}]{qi2023a}
\bibinfo{author}{Q.~Qi}, \bibinfo{author}{Q.~Bao}, \bibinfo{author}{A.~Y.
  Peng}, \bibinfo{author}{J.~Liu}, \bibinfo{author}{M.~Witbrock},
  \bibinfo{title}{A dynamic prompt-tuning method for data augmentation with
  associated knowledge}, \bibinfo{year}{2023}. \URLprefix
  \url{https://openreview.net/forum?id=hli7A0ioiS_}.
\bibitem[{Bao et~al.(2020)Bao, Ni, and Liu}]{10.1145/3373017.3373049}
\bibinfo{author}{Q.~Bao}, \bibinfo{author}{L.~Ni}, \bibinfo{author}{J.~Liu},
\newblock \bibinfo{title}{Hhh: An online medical chatbot system based on
  knowledge graph and hierarchical bi-directional attention},
\newblock in: \bibinfo{booktitle}{Proceedings of the Australasian Computer
  Science Week Multiconference}, ACSW '20, \bibinfo{publisher}{Association for
  Computing Machinery}, \bibinfo{address}{New York, NY, USA},
  \bibinfo{year}{2020}. \URLprefix
  \url{https://doi.org/10.1145/3373017.3373049}.
  \DOIprefix\doi{10.1145/3373017.3373049}.
\bibitem[{Ni et~al.(2022)Ni, Bao, Li, Qi, Denny, Warren, Witbrock, and
  Liu}]{Ni_Bao_Li_Qi_Denny_Warren_Witbrock_Liu_2022}
\bibinfo{author}{L.~Ni}, \bibinfo{author}{Q.~Bao}, \bibinfo{author}{X.~Li},
  \bibinfo{author}{Q.~Qi}, \bibinfo{author}{P.~Denny},
  \bibinfo{author}{J.~Warren}, \bibinfo{author}{M.~Witbrock},
  \bibinfo{author}{J.~Liu},
\newblock \bibinfo{title}{Deepqr: Neural-based quality ratings for
  learnersourced multiple-choice questions},
\newblock \bibinfo{journal}{Proceedings of the AAAI Conference on Artificial
  Intelligence} \bibinfo{volume}{36} (\bibinfo{year}{2022})
  \bibinfo{pages}{12826--12834}. \URLprefix
  \url{https://ojs.aaai.org/index.php/AAAI/article/view/21562}.
  \DOIprefix\doi{10.1609/aaai.v36i11.21562}.
\bibitem[{Pennington et~al.(2014)Pennington, Socher, and
  Manning}]{pennington2014glove}
\bibinfo{author}{J.~Pennington}, \bibinfo{author}{R.~Socher},
  \bibinfo{author}{C.~Manning},
\newblock \bibinfo{title}{{G}lo{V}e: Global vectors for word representation},
\newblock in: \bibinfo{booktitle}{EMNLP}, \bibinfo{address}{Doha, Qatar},
  \bibinfo{year}{2014}, pp. \bibinfo{pages}{1532--1543}.
\bibitem[{Cho et~al.(2014)Cho, van Merri{\"e}nboer, Bahdanau, and
  Bengio}]{cho2014properties}
\bibinfo{author}{K.~Cho}, \bibinfo{author}{B.~van Merri{\"e}nboer},
  \bibinfo{author}{D.~Bahdanau}, \bibinfo{author}{Y.~Bengio},
\newblock \bibinfo{title}{On the properties of neural machine translation:
  Encoder{--}decoder approaches},
\newblock in: \bibinfo{booktitle}{SSST-8}, \bibinfo{year}{2014}, pp.
  \bibinfo{pages}{103--111}.
\bibitem[{Xiong et~al.(2016)Xiong, Merity, and Socher}]{xiong2016dynamic}
\bibinfo{author}{C.~Xiong}, \bibinfo{author}{S.~Merity},
  \bibinfo{author}{R.~Socher},
\newblock \bibinfo{title}{Dynamic memory networks for visual and textual
  question answering},
\newblock in: \bibinfo{booktitle}{ICML}, \bibinfo{organization}{PMLR},
  \bibinfo{year}{2016}, pp. \bibinfo{pages}{2397--2406}.
\bibitem[{Gers et~al.(2000)Gers, Schmidhuber, and Cummins}]{gers2000learning}
\bibinfo{author}{F.~A. Gers}, \bibinfo{author}{J.~Schmidhuber},
  \bibinfo{author}{F.~Cummins},
\newblock \bibinfo{title}{Learning to forget: Continual prediction with lstm},
\newblock \bibinfo{journal}{Neural computation} \bibinfo{volume}{12}
  (\bibinfo{year}{2000}) \bibinfo{pages}{2451--2471}.
\bibitem[{Hochreiter and Schmidhuber(1997)}]{hochreiter1997long}
\bibinfo{author}{S.~Hochreiter}, \bibinfo{author}{J.~Schmidhuber},
\newblock \bibinfo{title}{Long short-term memory},
\newblock \bibinfo{journal}{Neural computation} \bibinfo{volume}{9}
  (\bibinfo{year}{1997}) \bibinfo{pages}{1735--1780}.
\bibitem[{Kumar et~al.(2016)Kumar, Irsoy, Ondruska, Iyyer, Bradbury, Gulrajani,
  Zhong, Paulus, and Socher}]{kumar2016ask}
\bibinfo{author}{A.~Kumar}, \bibinfo{author}{O.~Irsoy},
  \bibinfo{author}{P.~Ondruska}, \bibinfo{author}{M.~Iyyer},
  \bibinfo{author}{J.~Bradbury}, \bibinfo{author}{I.~Gulrajani},
  \bibinfo{author}{V.~Zhong}, \bibinfo{author}{R.~Paulus},
  \bibinfo{author}{R.~Socher},
\newblock \bibinfo{title}{Ask me anything: Dynamic memory networks for natural
  language processing},
\newblock in: \bibinfo{booktitle}{ICML}, volume~\bibinfo{volume}{48},
  \bibinfo{year}{2016}, pp. \bibinfo{pages}{1378--1387}.
\bibitem[{Hudson and Manning(2018)}]{hudson2018compositional}
\bibinfo{author}{D.~A. Hudson}, \bibinfo{author}{C.~D. Manning},
\newblock \bibinfo{title}{Compositional attention networks for machine
  reasoning},
\newblock in: \bibinfo{booktitle}{ICLR}, \bibinfo{year}{2018}.
\bibitem[{Kingma and Ba(2015)}]{kingma2014adam}
\bibinfo{author}{D.~P. Kingma}, \bibinfo{author}{J.~Ba},
\newblock \bibinfo{title}{Adam: {A} method for stochastic optimization},
\newblock in: \bibinfo{booktitle}{ICLR}, \bibinfo{year}{2015}.
\bibitem[{Ott et~al.(2019)Ott, Edunov, Baevski, Fan, Gross, Ng, Grangier, and
  Auli}]{ott2019fairseq}
\bibinfo{author}{M.~Ott}, \bibinfo{author}{S.~Edunov},
  \bibinfo{author}{A.~Baevski}, \bibinfo{author}{A.~Fan},
  \bibinfo{author}{S.~Gross}, \bibinfo{author}{N.~Ng},
  \bibinfo{author}{D.~Grangier}, \bibinfo{author}{M.~Auli},
\newblock \bibinfo{title}{fairseq: A fast, extensible toolkit for sequence
  modeling},
\newblock in: \bibinfo{booktitle}{NAACL-HLT}, \bibinfo{year}{2019}.

\end{thebibliography}
\section{Appendix}
\label{sec:appendix}
Following the closed-world assumption, we use the entities from the PARARULES which includes mainly PEOPLE and ANIMALS. We also consider negation, so there are 4 different scenarios in each depth for each category: PEOPLE with negation in the rules, PEOPLE without negation in the rules, ANIMALS with negation in the rules and ANIMALS without negation in the rules. All of the questions can be derived from the contexts and rules. If a question cannot be matched directly from the context, then it can be derived using rules. For the ANIMALS, we consider 14 different animal entities (``the bald eagle'', ``the tiger'', ``the bear'', ``the lion'', ``the wolf'', ``the crocodile'', ``the dinosaur'', ``the snake'', ``the leopard'', ``the cat'', ``the dog'', ``the mouse'', ``the rabbit'', ``the squirrel''), 7 different animals relationships (``is'', ``likes'', ``chases'', ``needs'', ``visits'', ``attacks'', ``sees''), and 28 different animals attributes (``big'', ``strong'', ``awful'', ``fierce'', ``heavy'', ``horrible'', ``powerful'', ``angry'', ``furry'', ``small'', ``cute'', ``lovely'', ``beautiful'', ``funny'', ``dull'', ``rough'', ``lazy'', ``slow'', ``sleepy'', ``boring'', ``tired'', ``reckless'', ``kind'', ``quiet'', ``round'', ``nice'', ``smart'', ``clever''). For the PEOPLE, we consider 9 different entities (``Anne'', ``Alan'', ``Bob'', ``Charlie'', ``Dave'', ``Erin'', ``Harry'', ``Gary'', ``Fiona''), 1 relationship (``is'') and 20 people attributes (``wealthy'', ``smart'', ``nice'', ``quiet'', ``kind'', ``poor'', ``dull'', ``rough'', ``bad'', ``sad'', ``short'', ``thin'', ``small'', ``little'', ``big'', ``strong'', ``high'', ``old'', ``young'', and ``huge'').

\begin{algorithm} 
	\caption{PARARULE-Plus data generation} 
	\label{PARARULE-Plus} 
	\begin{algorithmic}
		\REQUIRE Type of dataset: animal/people, add negation rule or not: Yes/No, \\
		reasoning depth $d \in {2,3,4,5}$, animal name list = ['the bald eagle', 'the tiger', 'the bear', \\'the lion', 'the wolf', 'the crocodile', 'the dinosaur', 'the snake', 'the leopard', 'the cat', \\'the dog', 'the mouse', 'the rabbit', 'the squirrel'], \\people name list = ['Anne', 'Alan', 'Bob', 'Charlie', 'Dave', 'Erin', 'Harry', 'Gary', 'Fiona'], \\animal relation list = ['is', 'is not', 'likes', 'chases', 'needs', 'visits', 'attacks', 'sees'],\\ people relation list = ['is', 'is not'],
		animal attribute list = ['kind', 'quiet', 'round', 'nice', 'smart', 'dull', 'rough', 'lazy', 'slow', 'sleepy', 'furry', 'small', 'cute', 'lovely', 'beautiful', \\'big', 'strong', 'awful', 'fierce', 'heavy'],
		people attribute list = ['big', 'strong', 'high',\\ 'huge', 'short', 'thin', 'small', 'little', 'wealthy', 'smart', 'nice', 'quiet', 'kind', 'poor', 'dull', 'rough', 'bad', 'sad', 'old', 'young']
		total\_list = []
		\FOR{reasoning depth $d$ in {2,3,4,5}}{
		\item item\_list = randomly select 4 animals/people name from animal/people name list
		\item \FOR{index in $range(0, len(item\_list))$}{
		\item item, item\_1, item\_2, item\_3 = item\_list[0], item\_list[1}, item\_list[2], item\_list[3]
		\item random shuffle animal/people relation list
		\IF{add negation rules == "No"}{\item context, question, label = load the template and fill item, item\_1, item\_2, item\_3 \\to the subject or object in the template, and we select one of the elements from\\ the animal/people relation list as the verb. For some facts, we add an item,\\ item\_1, item\_2, item\_3 as the subject, and one of the elements from the animal/\\people attribute list as the object to generate a dataset that only includes the \\depth = $d$, which means all questions have the same number of rules to derive \\the answer.}
		\ELSE{}{\item context, question, label = load the template, the template is similar to the above, \\but add a negation(not) in any rule to generate a dataset which only includes the depth = $d$}\ENDIF \\
		total\_list = total\_list.append({context, question, label, depth = $d$})
		  \ENDFOR}
		\ENDFOR
		\\
		return total\_list
	\end{algorithmic} 
\end{algorithm}




\end{document}